\documentclass[runningheads]{llncs}
\usepackage[T1]{fontenc}
\usepackage{graphicx}
\usepackage{color}
\usepackage{graphicx}
\usepackage{amsmath,amssymb} 
\usepackage{multirow}
\usepackage{graphicx,multirow}
\usepackage{wrapfig}
\usepackage{subfig}
\usepackage{amsmath}
\usepackage[table,xcdraw]{xcolor}
\usepackage{comment}
\usepackage{hyperref}
\usepackage{amsmath,amssymb} 
\usepackage{color}
\usepackage{tikz} 
\usepackage{pgfplots}
\usetikzlibrary{fit}
\pgfplotsset{compat=newest}%
\usepackage{enumitem}
\usepackage[accsupp]{axessibility}  
\begin{document}
\title{PS-ARM: An End-to-End Attention-aware Relation Mixer Network for Person Search}
%
%
\author{Mustansar Fiaz\inst{1}\orcidID{0000-0003-2289-2284} \and 
Hisham Cholakkal \inst{1} \and
Sanath Narayan \inst{2} \and
Rao Muhammad Anwer \inst{1} \and
Fahad Shahbaz Khan\inst{1}}
\authorrunning{F. Author et al.}
%
\institute{Department of computer Vision, Mohamed bin Zayed University of Artificial Intelligence, Abu Dhabi, UAE. \\
\email{(mustansar.fiaz, hisham.cholakkal, rao.anwer, fahad.khan)@mbzuai.ac.ae} \and 
Inception Institute of Artificial Intelligence, Abu Dhabi, UAE. }
\maketitle              
\begin{abstract}
Person search is a challenging problem with various real-world applications, that aims at joint person detection and re-identification of a query person from uncropped gallery images. Although, previous study focuses on rich feature information learning, it's still hard to retrieve the query person due to the occurrence of appearance deformations and background distractors. { In this paper, we propose a novel attention-aware relation mixer (ARM) module for person search, which exploits the global relation between different local regions within RoI of a person and make it robust against various appearance deformations and occlusion. The proposed ARM is composed of a  relation mixer block and a spatio-channel attention layer. The relation mixer block introduces a spatially attended spatial mixing and a channel-wise attended  channel mixing for effectively capturing  discriminative relation features within an RoI.  These discriminative relation features are further enriched  by introducing a spatio-channel attention where the  foreground and background discriminability is empowered in a joint spatio-channel space.}  Our ARM  module  is  generic  and it does not rely on  fine-grained  supervisions or topological assumptions, hence   being   easily  integrated  into   any  Faster  R-CNN based person search methods.  Comprehensive experiments  are performed on    two challenging  benchmark datasets: CUHK-SYSU  and PRW.  Our PS-ARM achieves state-of-the-art performance on both datasets.  On the challenging PRW dataset,  our PS-ARM achieves  an  absolute gain of 5\% in the  mAP score over SeqNet, while operating at a comparable speed. The source code and pre-trained models are available at  \href{https://github.com/mustansarfiaz/PS-ARM}(this https URL).

\end{abstract}
\section{Introduction}

Person search is a challenging computer vision problem where the task is to find a target query person in a gallery of whole scene images. The person search methods need to  perform pedestrian detection \cite{zhang2018occluded,liu2019high,pang2019mask} on the uncropped gallery images and do re-identification (re-id) \cite{Zheng_ReIDW_CVPR_2017,Liao_LMOR_CVPR_2015,Liao_LMOR_CVPR_2017} of the detected pedestrians.  In addition to  addressing the challenges associated with these individual  sub-tasks, both these tasks need to be simultaneously optimized within  person search. Despite numerous  real-world applications, person search is highly challenging due to the diverse nature of person detection and re-id sub-tasks  within the person search problem.  

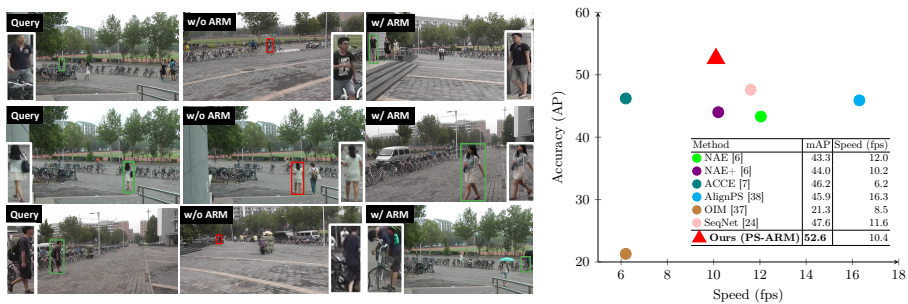
\begin{figure}[t!]
\centering
\resizebox{\textwidth}{!}{
\begin{minipage}{0.6\textwidth}
	\resizebox{7.90cm}{!}{%
	\includegraphics[width=\textwidth]{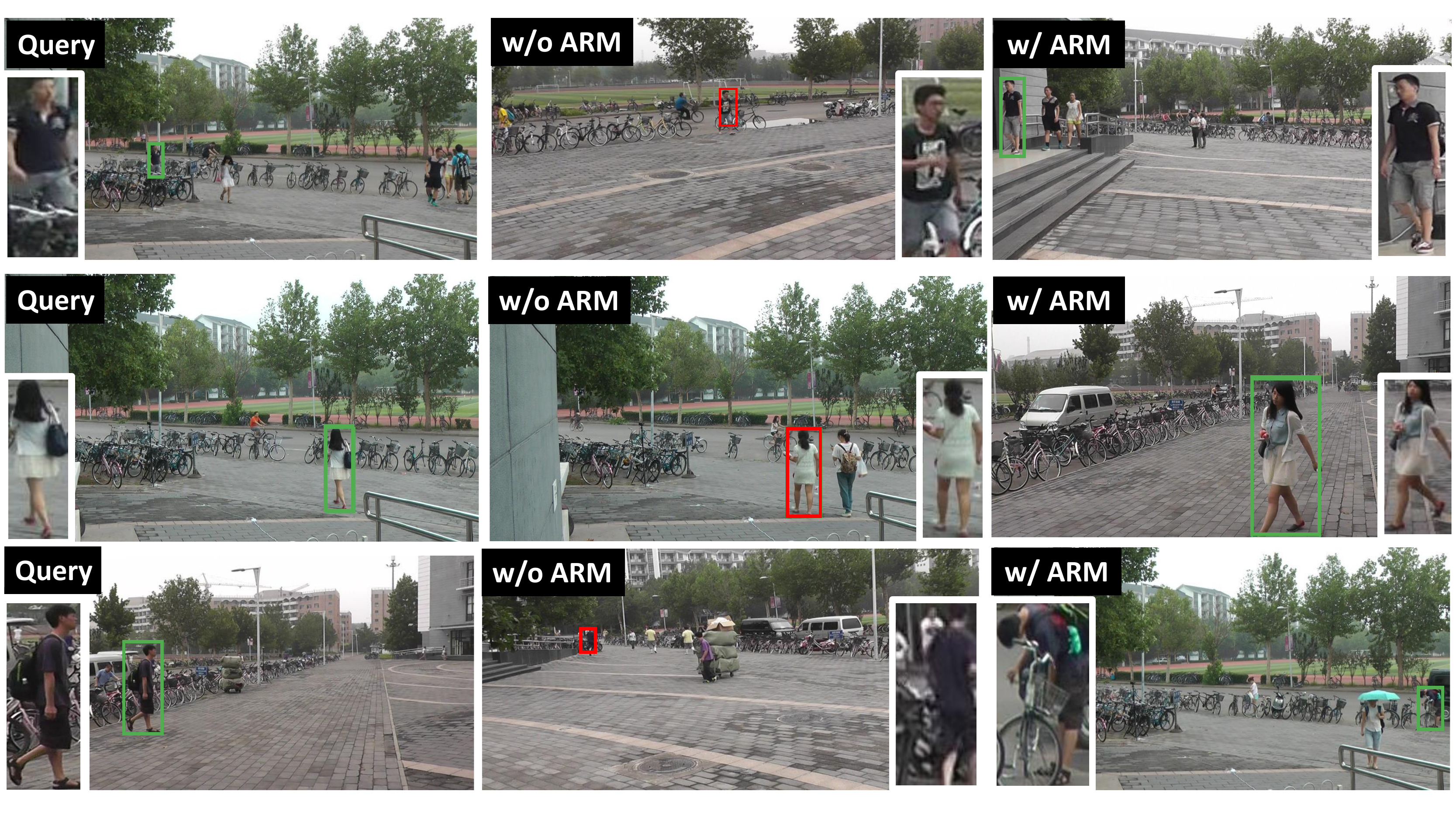}}
\end{minipage}
\qquad
\begin{minipage}{0.45\textwidth}
\centering
\resizebox{\textwidth}{!}{%
\begin{tikzpicture}
\begin{axis}[
axis lines = left,
ymin=20, ymax=60,
xmin=5, xmax=18,
xlabel= Speed (fps),
ylabel = Accuracy (AP),
]
\coordinate (legend) at (axis description cs:1.0,0.28);
\addplot[only marks,
mark=otimes*, violet,
mark size=3.5pt
]
coordinates {
(10.2,44)};\label{plot:NAE+}
\addplot[only marks,
mark=otimes*, green,
mark size=3.5pt
]
coordinates {
(12.04,43.3)};\label{plot:NAE}
\addplot[only marks,
mark=otimes*, pink,
mark size=3.5pt
]
coordinates {
(11.6,47.6)};\label{plot:SeqNet}
\addplot[only marks,
mark=otimes*, cyan,
mark size=3.5pt
]
coordinates {
(16.3,45.9)};\label{plot:AlignPS}
\addplot[only marks,
mark=otimes*, teal,
mark size=3.5pt
]
coordinates {
(6.2,46.2)};\label{plot:ACCE}
\addplot[only marks,
mark=otimes*, brown,
mark size=3.5pt
]
coordinates {
(6.2,21.3)};\label{plot:OIM}
\addplot[only marks,
mark=triangle*, red,
mark size=6.5pt
]
coordinates {
(10.1,52.6)};\label{plot:Ours}
\end{axis} 
\node[draw=none,fill=none, anchor=  east] at
(legend){\resizebox{4.7cm}{!}{
\begin{tabular}{l|r|r}
\hline
Method & mAP  & Speed (fps) \\ \hline
\ref{plot:NAE} NAE~\cite{chen2020norm} & 43.3  & 12.0  \\ 
\ref{plot:NAE+} NAE+~\cite{chen2020norm} & 44.0  & 10.2  \\  
\ref{plot:ACCE} ACCE~\cite{chen2021learning} & 46.2  & 6.2 \\  
\ref{plot:AlignPS} AlignPS~\cite{yan2021anchor} & 45.9  & 16.3 \\   
\ref{plot:OIM} OIM~\cite{xiao2017joint} & 21.3  & 8.5  \\  
\ref{plot:SeqNet} SeqNet~\cite{li2021sequential} & 47.6  & 11.6 \\  
\hline 
\ref{plot:Ours} \textbf{Ours (PS-ARM)} & \textbf{52.6 } & 10.4 \\ \hline  
\end{tabular} }};
\end{tikzpicture}
}
\end{minipage}}
 \caption{\textbf{On the left:}  Qualitative comparison showing different query examples and their corresponding top-1 matching results obtained with \textit{and} without our ARM module in the same  base framework. Here, true and false matching results are marked  in green and red,  respectively. These examples depict appearance deformations and distracting backgrounds in the gallery images for the query person. Our ARM module that explicitly captures discriminative relation features better handle the appearance deformations in these examples. 
\textbf{On the right:}
Accuracy (AP) vs.~speed (frames per second) comparison with state-of-the-art  person search methods on PRW test set. All methods are reported with a  Resnet50 backbone and speed is computed over V100 GPU. Our approach (PS-ARM) achieves an absolute mAP gain of 5\%  over SeqNet while operating at a comparable speed.
}
\label{intro_fig_PS}
\end{figure}

Person search approaches can be broadly grouped into two-step \cite{Zheng_PRW_CVPR_2017,Chen_MGTS_ECCV_2018,Han_RDLR_ICCV_2019} and one-step  methods\cite{xiao2017joint,yan2019learning,chen2020norm}. In two-step approaches, person detection and re-id are performed separately using two different steps. In the first step  a detection network such as Faster R-CNN is employed to detect pedestrians. In the second step detected persons are  first cropped and re-sized from the \textit{input image}, then utilized in another  independent network for the re-identification of the cropped pedestrians.  Although two-step methods provide promising results, they are computationally expensive. 
Different to two-step methods, one step methods employ a unified framework where the  backbone networks are shared  for the  detection and identifications of persons. For a given uncropped image, one-step methods  predict the box coordinates and re-id features for all persons in that image. 
One-step person search approaches such as  \cite{chen2020norm,li2021sequential}  generally extend Faster R-CNN  object detection frameworks  by introducing an additional branch  to produce re-id feature embedding, and the whole network is jointly  trained end-to-end.  Such  methods often struggle while the target person  in the galley images has large  appearance deformations such as  pose variation, occlusion, and overlapping background distractions within the region of interest (RoI) of a target person (see Figure.~\ref{intro_fig_PS}). 
\subsection{Motivation}
To motivate our approach, we first distinguish two desirable characteristics to be considered when designing a Faster R-CNN based person search framework that is robust to appearance deformations (e.g. pose variations, occlusions) and  background distractions occurring in the query person (see Figure.~\ref{intro_fig_PS}).\\
\noindent\textit{Discriminative Relation Features through Local Information Mixing:}
The position of different local person regions within an RoI can vary in case of appearance deformations such as pose variations and occlusions. This is likely to deteriorate the quality of re-id features, leading to inaccurate person matching. Therefore, a dedicated mechanism is desired that generates discriminative relation features by globally mixing relevant information from different local regions within an RoI.  
To ensure a straightforward integration into existing person search pipelines, such a mechanism is further expected to learn discriminative relation features without requiring fine-level region supervision or topological body approximations. 


{\noindent\textit{Foreground-Background Discriminability for Accurate Local Information Mixing:}
The quality of the aforementioned relation features rely on the assumption that the RoI region only contains foreground (person) information. However, in real-world scenarios the RoI regions are likely to contain unwanted background information due to less accurate bounding-box locations. Therefore, discriminability of the foreground from the background is essential for accurate local information mixing to obtain discriminative relation features. Further, such a FG/BG discrimination is expected to also improve the detection performance. }

\subsection{Contribution}
We propose a novel end-to-end one-step  person search method with the following novel contributions.
    We propose a novel attention-aware relation mixer (ARM) module that strives to capture global relation between different local person regions through global mixing of local information while simultaneously suppressing background distractions within an RoI. 
    Our ARM module comprises a relation mixer block  and a  spatio-channel attention layer. The relation mixer block captures discriminative relation features through a spatially-attended spatial mixing and a channel-wise attended channel mixing. These discriminative relation features are further enriched by  the spatio-channel attention layer performing {foreground/background discrimination in a joint spatio-channel space}.  
    Comprehensive experiments are performed on  two challenging  benchmark datasets: CUHK-SYSU \cite{xiao2017joint} and PRW \cite{zheng2017person}.  On both datasets, our PS-ARM performs favourably against state-of-the-art approaches. On the challenging PRW benchmark,  our PS-ARM achieves a mAP score of 52.6\%.  Our ARM  module  is  generic  and  can  be  easily  integrated  to  any  Faster  R-CNN based person search methods. 
   Our PS-ARM   provides an absolute gain of 5\%  mAP score over SeqNet, while operating at a comparable  speed  (see Figure. ~\ref{intro_fig_PS}), resulting in a mAP score of 52.6\% on the challenging PRW dataset. 
\section{Related Work}
Person search is a challenging computer vision problem with numerous real-world applications.  As mentioned earlier, existing person methods can be broadly 
classified into two-step and one-step methods. Most existing two-step person search approaches address this problem by first detecting the pedestrians, followed by  cropping and resizing  into a fixed resolution before passing to the re-id network that identifies the  cropped  pedestrian \cite{zheng2017person,chen2018person,han2019re,dong2020instance,lan2018person}.  These methods generally employ two different backbone networks for the detection and re-identifcation.   

On the other hand, several one-step person search methods  employ feature pooling strategies such as, RoIPooling or RoIAlign pooling to obtain a scale-invariant  representation for the re-id sub-task.  
\cite{chen2018person} proposed a two-step method to learn  robust person features by exploiting person foreground maps using pretrained segmentation network. Han et al. \cite{han2019re}  introduced a bounding box refinement mechanism for person localization.  Dong et al. \cite{dong2020instance} utilized the similarity between the query and query-like features to reduce the number of proposals for re-identification.  Zhang et al. \cite{zheng2017person} introduced the challenging  PRW dataset.  A multi-scale feature pyramid was introduced in \cite{lan2018person}  for improving person search under scale variations. Wang et al. \cite{wang2020tcts} proposed a method to address the  inconsistency between the detection and re-id  sub-tasks.

Most one-step person search methods ~\cite{xiao2017joint,xiao2019ian,liu2017neural,chang2018rcaa,yan2019learning,dong2020bi,chen2020norm,munjal2019query,han2021end,li2021sequential} are developed based on Faster R-CNN object detector \cite{ren2015faster}.   These methods generally  introduce an additional branch to Faster R-CNN and jointly  address  the  detection and Re-ID subtasks.
One of the earliest Faster R-CNN based  one-step person approach is  \cite{xiao2017joint}, which proposed an online instance matching (OIM) loss.   Xiao et al. \cite{xiao2019ian} introduced a  center loss to explore intra-class compactness. For generating  person proposals, Liu et al. \cite{liu2017neural} introduced a mechanism to iteratively shrink the search area based on query guidance. Similarly, Chang et al. \cite{chang2018rcaa} used reinforcement learning to address the person search problem. Chang et al. \cite{yan2019learning} exploited complementary cues based on graph learning framework.    Dont et al. \cite{dong2020bi}
proposed Siamese based Bi-directional Interaction Network (BINet) to mitigate redundant context information outside the BBoxes.
On the contrary, Chen et al. \cite{chen2020norm} proposed Norm Aware Embedding (NAE) to alleviate the conflict between person localization and re-identification by computing magnitude and angle of the embedded features respectively. 

Chen at al. \cite{chen2020hierarchical} developed a Hierarchical Online Instance Matching loss to guide the feature learning by exploiting the hierarchical relationship between detection and re-identification. A  query-guided proposal network (QGPN) is proposed by Munjal et al. \cite{munjal2019query} to learn query guided re-identification score. 
H Li et al. \cite{li2021sequential} proposed a Sequential End-to-end Network (SeqNet) to refine the proposals by introducing  Faster R-CNN as a proposal generator into the NAE pipeline to get refined features for detection and re-identification. The  Faster R-CNN based one-step person search approaches often struggle while the target undergoes large appearance deformations or come across with  distracting background objects within RoI.  
To address this,  we propose a novel person search method, PS-ARM, where a novel ARM module is introduced to capture global relation between different local regions within an RoI. Our PS-ARM  enables  accurate  detection and re-identification  of person instances under 
under challenging scenarios such as pose variation and distracting backgrounds (See Figure.~\ref{intro_fig_PS}).

\section{Method}

\begin{figure}[t!]
\includegraphics[width=\textwidth]{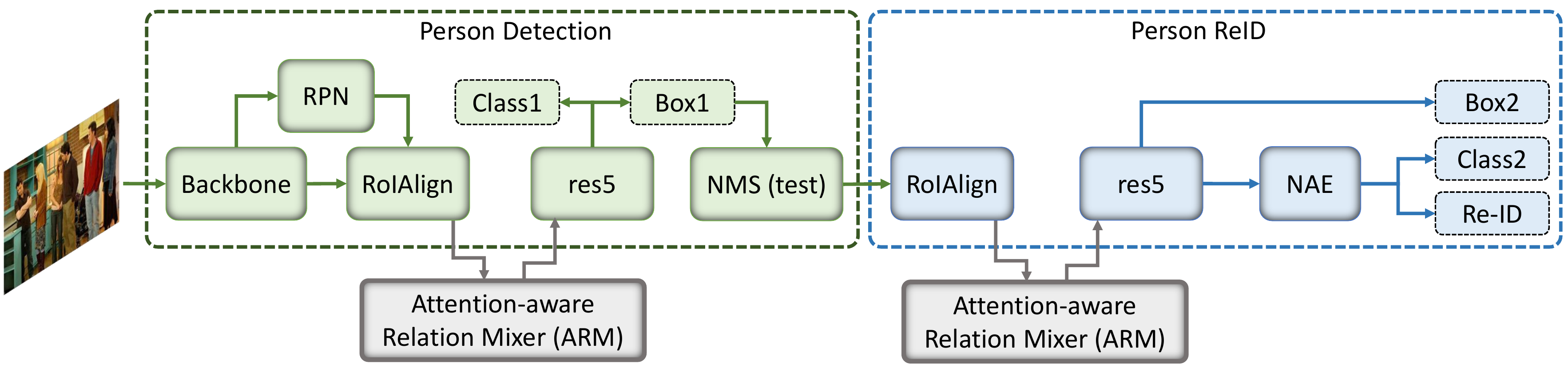}
\caption{The overall architecture of the proposed PS-ARM framework. It comprises a person  detection branch  (shown in green) and  person re-ID branch (shown in blue).  { The  person detection branch predicts the initial box locations whereas the  person re-id branch  refines the box locations and perform a   norm-aware embedding  (NAE) to disentangle the  detection and re-id.} The focus of our design is the introduction of a novel \textit{attention-aware relation Mixer} (ARM) module (shown in grey) to the detection and re-id branches. Our ARM module enriches standard RoI Align pooled features by capturing discriminative relation features between different local regions within an RoI. The resulting enriched features are used for box regression and classification in the detection branch, whereas these features are used to refine the box locations, along with generating a  norm-aware embedding  for box classification (person vs background) and re-id feature prediction in the re-id branch.} 
\label{proposed_framework}
\end{figure}

\subsection{Overall Architecture}
Figure.~\ref{proposed_framework} shows overall architecture of the proposed framework. It comprises  
a person detection branch (shown in green) followed by a  person re-ID branch (shown in blue).  The  person detection branch follows the structure of standard Faster R-CNN, which comprises a ResNet  backbone ($res1$-$res4$), a region proposal network (RPN), RoIAlign pooling, and a prediction head for box regression and classification. The person re-id branch takes the boxes predicted by the person detection branch as input and performs RoIAlign pooling on these predicted box locations.  { The resulting RoI Align pooled features are utilized  to perform re-identification.
We adopt norm-aware embedding (NAE) that is designed to separate the detection and re-identification using shared feature representation.}
During inference,  the person re-id branch   takes only unique boxes (obtained by non-maximum suppression algorithm)  from the person detection branch and performs a context bipartite graph matching for the re-id similar to \cite{li2021sequential}. The above-mentioned standard detection and re-id branches serve as a base network to which we introduce our novel attention-aware relation mixer (ARM) module that enriches the  RoI features for accurate person search.

The focus of our design is the introduction of a novel ARM module (shown in grey).  Specifically, we integrate our ARM module between the RoIAlign and convolution blocks ($res5$) in both the person detection and re-id branches of the base framework, without  sharing the parameters between both branches. Our proposed ARM  module strives to  enrich  standard  RoI  Align pooled features by capturing discriminative relation features between different local regions within an RoI through global mixing of local information.  To ensure effective enrichment of RoI Align pooled features,  {we further introduce a foreground/background discrimination mechanism in our ARM module. Our ARM module strives to simultaneously improve both detection and re-id sub-tasks. Therefore, the output is passed to norm-aware embedding to decouple the features for the contradictory detection and re-id tasks. }
Furthermore, our ARM module is generic and can be easily integrated to other Faster R-CNN based person search methods. Next, we present the details of the proposed ARM module.

\begin{wrapfigure}{r}{0.5\textwidth}
 \vspace{-0.3cm}
  \begin{center}
    \includegraphics[width=0.48\textwidth]{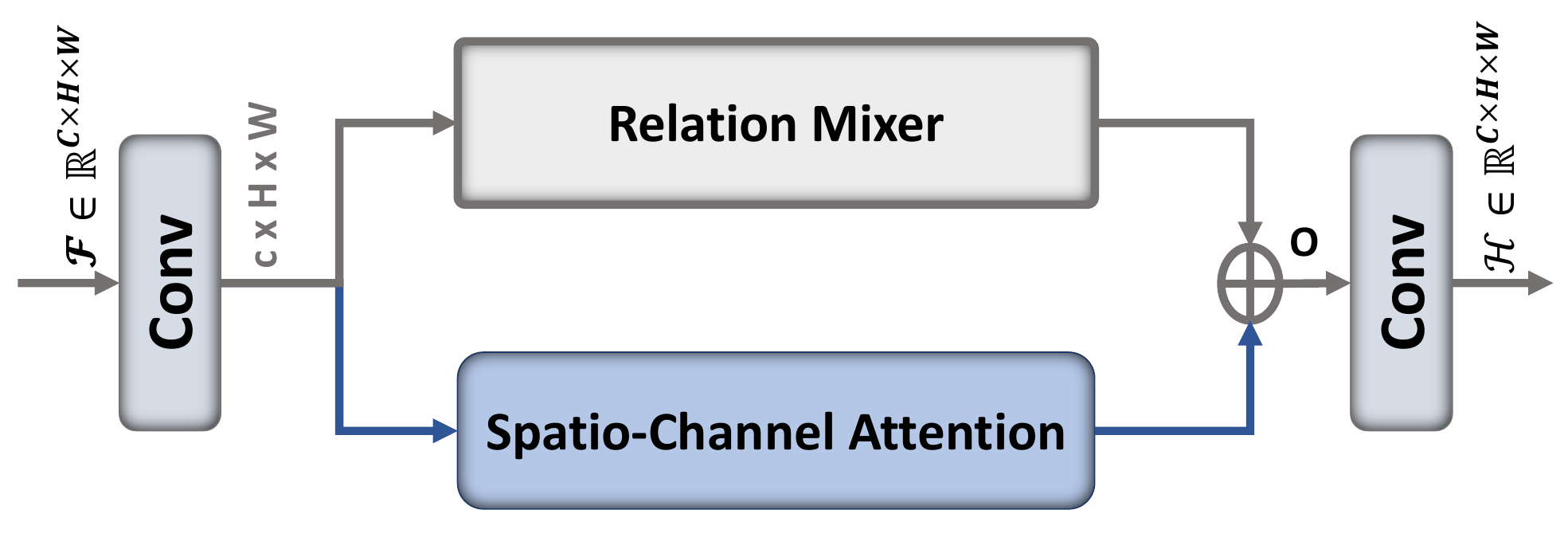}
  \end{center}
  \caption{ The network structure of our ARM module. The module takes RoI Align pooled features as input and captures inter-dependency between different local regions, while simultaneously suppressing  background distractions for the person search problem. To achieve this objective, ARM module comprises a relation mixer block and a joint  spatio-channel attention layer.}
  \vspace{-0.3cm}
   \label{proposed_ARM}
\end{wrapfigure}

\subsection{Attention-aware Relation Mixer (ARM) Module}

Our ARM module is shown in Figure.~\ref{proposed_ARM}. It comprises a relation mixer block and a  spatio-channel attention layer.  Our relation mixer block captures relation between different sub-regions (local regions) within an RoI. The resulting features are further enriched by our spatio-channel attention  that attend to  relevant input features in a  joint spatio-channel space. Our ARM module takes RoIAlign pooled feature  $\mathcal{F} \in \mathbb{R}^{C\times H\times W}$ as input. Here, $H$, $W$, $C$ are the height,  width  and  number of channels of the RoI  feature. For computational efficiency, the number of channels are reduced to $c=C/4$ through a point ($1\times1$) convolution  layer before passing to relation mixer and spatio-channel attention blocks. \\

\noindent\textbf{Relation Mixer Block:}
As mentioned earlier, our relation mixer block is introduced to capture the relation between different sub-regions (local regions) within an RoI. This is motivated by the fact that the local regions of a person share certain `standard' prior relationships among local regions,  across RoIs of different person and it is desirable to explicitly learn these inter-dependencies  without any supervision.  
One such module that can learn/encode such inter-dependencies, is the MLP-mixer \cite{tolstikhin2021mlp} that performs spatial `token' mixing followed by `pointwise' feature refinement.
Compared to other context aggregators  \cite{kaiser2017depthwise,Vaswani_Att_NIPS_2017,kipf2016semi}, the mlp-mixer is more static, dense, and does not share parameters \cite{gao2021container}. The core operation of the MLP-Mixer is  transposed affinity matrix  on a single feature group, which computes  the affinity matrix with non-sharing $W_{MLP-1}$ parameters as: $\mathcal{A} = (W_{MLP-1})^T$. To this end, MLP mixer  contains  a spatial mixer and a channel mixer.
The spatial mixer comprise of a layer norm,  skip connection and a token-mixing MLP with two fully-connected layers and a GELU nonlinearity. Similarly, the channel mixer employs a channel-mixing MLP, layer norm, skip connection and dropout.
The MLP mixer conceptually acts as a persistent relationship memory that can learn and encode the prior relationships among the local regions of an object at a global level. 
 {To this end, we introduce our relation mixer comprising a spatially attended spatial mixer and a channel-wise attended  channel mixer. our ARM module with residual connection not only enabled using MLP mixer for the first time in the problem of person search, but also provided impressive performance gain over the base framework.}
\begin{figure}[t!]
\centering
\includegraphics[width=\textwidth]{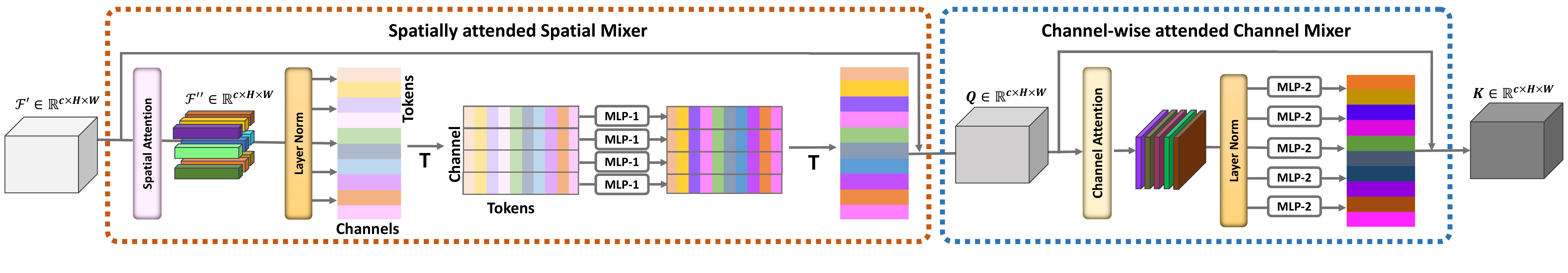}
\caption{Structure of relation mixer block within our ARM module. It comprises a spatially attended spatial mixing operation where important local spatial regions will be emphasized using a spatial attention before globally mixing them  across all spatial regions (tokens) within each channel  using  MLP-1 shared across all channels. Following this spatial mixing, we perform a channel attention to emphasize informative channels before globally mixing the channels for each local spatial region (token) using  MLP-2 shared across all spatial regions. 
}
\label{relation_mixer_architecture}
\end{figure}\\

\noindent\textit{\textbf{Spatially attended Spatial Mixer:}} {While learning the inter-dependencies of local RoI sub-regions using standard MLP mixer, the background regions are likely to get entangled with the foreground regions, thereby adversely affecting the resulting feature embedding used for the re-id and box predictions. In order to discriminate the irrelevant background information at token level, we introduce a spatial attention before performing  token (spatial) mixing within our MLP mixer for emphasizing the foreground regions. } In our spatial attention, we employ   pooling operations along the channel axis, followed by    convolution and sigmoid layers to generate a 2D spatial attention weights  ${M_s} \in \mathbb{R}^{1\times H\times W}$. These attention weights  are broadcasted along the channel dimension to generate the spatial attention ${M'_s} \in \mathbb{R}^{c\times H\times W}$. 
For a given feature  $\mathcal{F'} \in \mathbb{R}^{c\times H\times W}$, we obtain the spatially attended feature map 
$\mathcal{F''}=\mathcal{F'}\odot {M}'_s$. Here $\odot$  denotes element-wise multiplication. 
{These spatially attended features ($\mathcal{F''}$)  are expected to  discriminate irrelevant (background)  spatial regions from the foreground. }These features are ($\mathcal{F''}$)  input to a shared multi-layer perceptron (MLP-1) for globally mixing local features (within $\mathcal{F''}$)  across all spatial regions (tokens). Our spatially attended spatial mixing   strives to  achieve accurate  spatial mixing  and outputs the feature map Q (see Figure.~\ref{relation_mixer_architecture}).

\noindent\textbf{\textit{Channel-wise attended Channel Mixer:}} 
To further prioritize the feature channels of $Q$ that are relevant for detection and re-id  of person instances, we introduce a channel attention  before  channel mixing.    
 Our  channel attention weights ${M_c} \in \mathbb{R}^{c\times  1}$ are generated through spatial pooling, fully connected (fc) and sigmoid layers, which are broadcasted along  the spatial dimension to generate the channel attention weights ${M'_c} \in \mathbb{R}^{c\times H\times W}$.    
Similar to spatial attention, these channel weights are element-wise multiplied with the  feature map to obtain channel-wise attended feature map. The resulting features are expected to   emphasize only the  channels that are relevant  for effective channel-mixing within our relation mixing block.  {Our channel mixing employs another shared  MLP (MLP-2) for  global mixing of channel information. The final output of our relation mixer block results is feature maps $K\in \mathbb{R}^{c\times H\times W}$.}

\noindent\textbf{Spatio-channel Attention Layer:}
Our relation mixer block performs the mixing operations by treating the spatial and channel  information in a disjoint manner. But, in many scenarios, all spatial regions within a channel and all channels at a given spatial location are not equally informative.  Hence, it is desired to treat the entire spatio-channel information  as a joint space.  With this objective, we introduce a  joint spatio-channel attention layer within our ARM module to further  { improve  the foreground/background discriminability of RoIAlign pooled features.}
Our spatio-channel attention  layer utilizes parameter-free 3D attention weights obtained based on \cite{yang2021simam} to modulate the 3D spatio-channel RoI pooled features. These spatio-channel attended features are aggregated  with the  relation mixer output to produce enriched features $O$  for the person search task. These enriched features projected back to $C$ channels ($\mathcal{H} \in \mathbb{R}^{C\times H\times W}$) and taken as input to the $res5$ block. 

 {In summary, within our ARM module, the relation mixer targets the global relation between different local regions within RoI  and captures the discriminative relation features in  disjoint spatial and channel spaces. The resulting features are further enriched by  a spatio-channel attention that performs foreground/background discrimination in a joint spatio-channel space. }
\subsection{Training and Inference}
For training and inference, we follow a strategy similar to \cite{chen2020norm,li2021sequential}. 
Our PS-ARM is trained end-to-end  with a  loss formulation similar to  \cite{li2021sequential}.
That is, in the person detection branch, similar to Faster R-CNN, we employ Smooth-L1 and cross entropy losses for  box regression and classifications.  For the person re-id branch, 
we employ three additional loss terms similar to  \cite{chen2020norm} for regression, classification and re-ID.  Both these branches are trained by utilizing an IoU threshold  of 0.5 for selecting  positive and negative samples.  

During inference, we first obtain the re-id feature for a given query by using the provided bounding box. Then, for the gallery images,  the predicted boxes and their  re-id features are obtained from the re-id branch. Finally, employ cosine similarity between the re-id features  to  match a query person with an arbitrarily  detected person in the galley.  

\section{Experiments}
We perform the experiments on two person search datasets (\textit{i.e.,} CUHK-SYSU \cite{xiao2017joint}) and PRW \cite{zheng2017person} to demonstrate the effectiveness of our PS-ARM and compare it with the state-of-the-art methods.

\subsection{Dataset and Evaluation Protocols}
\noindent\textbf{CUHK-SYSU~\cite{xiao2017joint}:} is a large scale person search dataset with 96,143 person bounding boxes from a total of 18,184 images. The training and testing sets contains 11,206 images, 55,272 pedestrians, and 5,532 identities and test set includes 6,978 images, 40,871 pedestrians, and 2,900 identities. Instead of using full gallery during inference, different gallery sizes are used for each query from 50 to 4000. The default gallery size is set to 100.

\noindent\textbf{PRW~\cite{zheng2017person}:} is composed of video frames recorded by six cameras that are being installed at different location in Tsinghua University. The dataset has a total 11,816 frames containing 43,110 person bounding boxes.
In training set, 5,704 images are annotated with 482 identities. The test set has 2,057 frames are labelled as query persons while gallery set has 6,112 images. Hence, the gallery size of PRW dataset is notably larger compared to CUHK-SYSU gallery set.

\noindent\textbf{Evaluation Protocol:} We follow two standard protocol for person search peformance evaluation of mean Average Precision (mAP) and top-1 accuracy. The mAP is computed by averaging over all queries with an intersection-over-union (IoU) threshold of 0.5. The top-1 accuracy is measured according to the IoU overlaps between the top-1 prediction and ground-truth with the threshold value set to 0.5. 

\noindent\textbf{Implementation Details:}
We used ResNet-50 as our backbone network. We followed \cite{li2021sequential} and utilized Stochastic Gradient Descent (SGD), set momentum and decay to 0.9 and $5 \times 10^{-4}$, respectively.  We trained the model for 12 epochs over CUHK-SYSU dataset PRW dataset.  During training, we used the batch-size of 3 with input size $900 \times 1500$ and set initial learning rate to 0.003 which is warmed up at first epoch and decayed by $0.1$ at 8th epoch. During inference, the NMS threshold value is set to 0.4. The code is implemented in PyTorch \cite{paszke2019pytorch}. The code and trained model will be publicly released.  

\subsection{Comparison with State-of-the-art Methods}
Here, we compare our approach with state-of-the-art one-step and two-step person search methods in literature on  two datasets: CUSK-SYSU and PRW. 

\begin{table}[t!]
\begin{center}
\caption{ State-of-the-art comparison on CUHK and PRW test sets in terms of mAP and top-1 accuracy.  On both datasets, our PS-ARM performs favourably against existing approaches.  All the methods here utilize the same ResNet50 backbone. When compared with recently introduced SeqNet, our PS-ARM provides an absolute mAP gain of 5\% on the challenging PRW dataset. Also, introducing our novel ARM module to a popular Faster R-CNN based approach (NAE \cite{chen2020norm}), provides an absolute mAP gain of 3.6\%.   }
\label{tbl:quantitative}
\scalebox{0.9}{
\begin{tabular}{|cl|cc|cc|c|}
\hline
\multicolumn{2}{|c|}{}  & \multicolumn{2}{|c|}{CUHK-SYSU}   & \multicolumn{2}{|c|}{PRW}    \\ \cline{3-6}
\multicolumn{2}{|c|}{\multirow{-2}{*}{Method}} & \multicolumn{1}{|c|}{mAP} & \multicolumn{1}{|c|}{top-1} & \multicolumn{1}{|c|}{mAP} & \multicolumn{1}{|c|}{top-1} \\ \cline{3-6} \hline \hline
\multicolumn{1}{|c|}{  \multirow{6}{*}{\rotatebox[origin=c]{90}{Two-step}}}  & CLSA \cite{lan2018person} & \multicolumn{1}{c|}{87.2} & 88.5& \multicolumn{1}{c|}{38.7} & 65.0  \\
\multicolumn{1}{|c|}{} & IGPN \cite{dong2020instance}   & \multicolumn{1}{c|}{90.3}  & 91.4 & \multicolumn{1}{c|}{42.9} & 70.2   \\ 
\multicolumn{1}{|c|}{} & RDLR \cite{han2019re}   & \multicolumn{1}{c|}{93.0}     &  94.2 & \multicolumn{1}{c|}{42.9}     &   70.2      \\ 
\multicolumn{1}{|c|}{} & MGTS  \cite{chen2018person} & \multicolumn{1}{c|}{83.0}   &83.7 & \multicolumn{1}{c|}{32.6}  &   72.1     \\
\multicolumn{1}{|c|}{} & MGN+OR \cite{yao2020joint}  & \multicolumn{1}{c|}{93.2}   & 93.8 & \multicolumn{1}{c|}{52.3}  &  71.5       \\
\multicolumn{1}{|c|}{} & TCTS \cite{wang2020tcts}  & \multicolumn{1}{c|}{93.9}  & 95.1 & \multicolumn{1}{c|}{46.8}     &  87.5   \\ \hline
\multicolumn{1}{|c|}{  \multirow{22}{*}{\rotatebox[origin=c]{90}{End-to-end}}} & OIM  \cite{xiao2017joint} & \multicolumn{1}{c|}{75.5} & 78.7  & \multicolumn{1}{c|}{21.3} & 49.9 \\
\multicolumn{1}{|c|}{} & RCAA \cite{chang2018rcaa}   & \multicolumn{1}{c|}{79.3} & 81.3   & \multicolumn{1}{c|}{-}  & -       \\ 
\multicolumn{1}{|c|}{} & NPSM \cite{liu2017neural}   & \multicolumn{1}{c|}{77.9}     & 81.2    & \multicolumn{1}{c|}{24.2}     &  53.1        \\
\multicolumn{1}{|c|}{} & IAN \cite{xiao2019ian}   & \multicolumn{1}{c|}{76.3}     & 80.1    & \multicolumn{1}{c|}{23.0}     & 61.9     \\
\multicolumn{1}{|c|}{} & QEEPS \cite{munjal2019query}   & \multicolumn{1}{c|}{88.9}     & 89.1    & \multicolumn{1}{c|}{37.1}     & 76.7      \\
\multicolumn{1}{|c|}{} & CTXGraph \cite{yan2019learning}   & \multicolumn{1}{c|}{84.1}     & 86.5    & \multicolumn{1}{c|}{33.4}     &73.6      \\
\multicolumn{1}{|c|}{} & HOIM \cite{chen2020hierarchical}   & \multicolumn{1}{c|}{89.7}     & 90.8   & \multicolumn{1}{c|}{39.8}     &  80.4     \\
\multicolumn{1}{|c|}{} & BINet \cite{dong2020bi}   & \multicolumn{1}{c|}{90.0}     & 90.7   & \multicolumn{1}{c|}{45.3}     & 81.7   \\
\multicolumn{1}{|c|}{} & AlignPS \cite{yan2021anchor}   & \multicolumn{1}{c|}{93.1}     & 96.4   & \multicolumn{1}{c|}{45.9}     & 81.9      \\
\multicolumn{1}{|c|}{} & PGSFL \cite{kim2021prototype}   & \multicolumn{1}{c|}{92.3}     & 94.7   & \multicolumn{1}{c|}{44.2}     & 85.2    \\
\multicolumn{1}{|c|}{} & DKD \cite{zhang2021diverse}   & \multicolumn{1}{c|}{93.1}     & 94.2  & \multicolumn{1}{c|}{50.5}     & 87.1    \\
\multicolumn{1}{|c|}{} & NAE+ \cite{chen2020norm}   & \multicolumn{1}{c|}{92.1}     & 94.7   & \multicolumn{1}{c|}{44.0}     & 81.1     \\
\multicolumn{1}{|c|}{} & PBNet \cite{tian2020end}   & \multicolumn{1}{c|}{90.5}     & 88.4   & \multicolumn{1}{c|}{48.5}     & 87.9     \\
\multicolumn{1}{|c|}{} & DIOIM \cite{dai2020dynamic}   & \multicolumn{1}{c|}{88.7}     & 89.6   & \multicolumn{1}{c|}{36.0}     & 76.1    \\
\multicolumn{1}{|c|}{} & APNet \cite{zhong2020robust}   & \multicolumn{1}{c|}{88.9}     & 89.3   & \multicolumn{1}{c|}{41.2}     & 81.4     \\
\multicolumn{1}{|c|}{} & DMRN \cite{han2021decoupled}   & \multicolumn{1}{c|}{93.2}     & 94.2   & \multicolumn{1}{c|}{46.9}     & 83.3   \\
\multicolumn{1}{|c|}{} & CAUCPS \cite{han2021context}   & \multicolumn{1}{c|}{81.1}     & 83.2 & \multicolumn{1}{c|}{41.7} & 86.0 \\
\multicolumn{1}{|c|}{} &  ACCE \cite{chen2021learning}   & \multicolumn{1}{c|}{93.9} & 94.7 & \multicolumn{1}{c|}{46.2} & 86.1 \\
\multicolumn{1}{|c|}{} & NAE \cite{chen2020norm}   & \multicolumn{1}{c|}{91.5}     & 92.4   & \multicolumn{1}{c|}{43.3}     & 80.9    \\
\multicolumn{1}{|c|}{} & SeqNet  \cite{li2021sequential}   & \multicolumn{1}{c|}{94.8} & 95.7   & \multicolumn{1}{c|}{47.6}     & 87.6     \\
\cline{2-6}

\multicolumn{1}{|c|}{} &\textbf{ Ours (NAE + ARM})   & \multicolumn{1}{c|}{93.4}  & 94.2   & \multicolumn{1}{c|}{46.9} &   81.4 \\

\multicolumn{1}{|c|}{} &  \textbf{ Ours (PS-ARM)}    & \multicolumn{1}{c|}{\textbf{95.2}} & \textbf{ 96.1} & \multicolumn{1}{c|}{\textbf{52.6}}  & \textbf{88.1} \\
\multicolumn{1}{|c|}{} &  \textbf{ Ours (Cascaded PS-ARM)}    & \multicolumn{1}{c|}{\textbf{-}} & \textbf{-} & \multicolumn{1}{c|}{\textbf{53.1}}  & \textbf{88.3} \\
\hline
\end{tabular}}
\end{center}
\end{table}

\noindent\textbf{CUHK-SYSU Comparison}: Table.~\ref{tbl:quantitative} shows the comparison of our PS-ARM with state-of-the-art two-step and single-step end-to-end methods with the gallery size of 100. Among existing two-step methods,  MGN+OR \cite{yao2020joint} and TCTS \cite{wang2020tcts} achieves mAP of 93.2 and 93.9, respectively. Among existing single-step end-to-end methods,  SeqNet \cite{li2021sequential} and  AlignPS \cite{yan2021anchor} obtains mAP of  94.8\%, 93.1\%  respectively.


\begin{wrapfigure}{r}{0.5\textwidth}
  \begin{center}
    \includegraphics[width=0.48\textwidth]{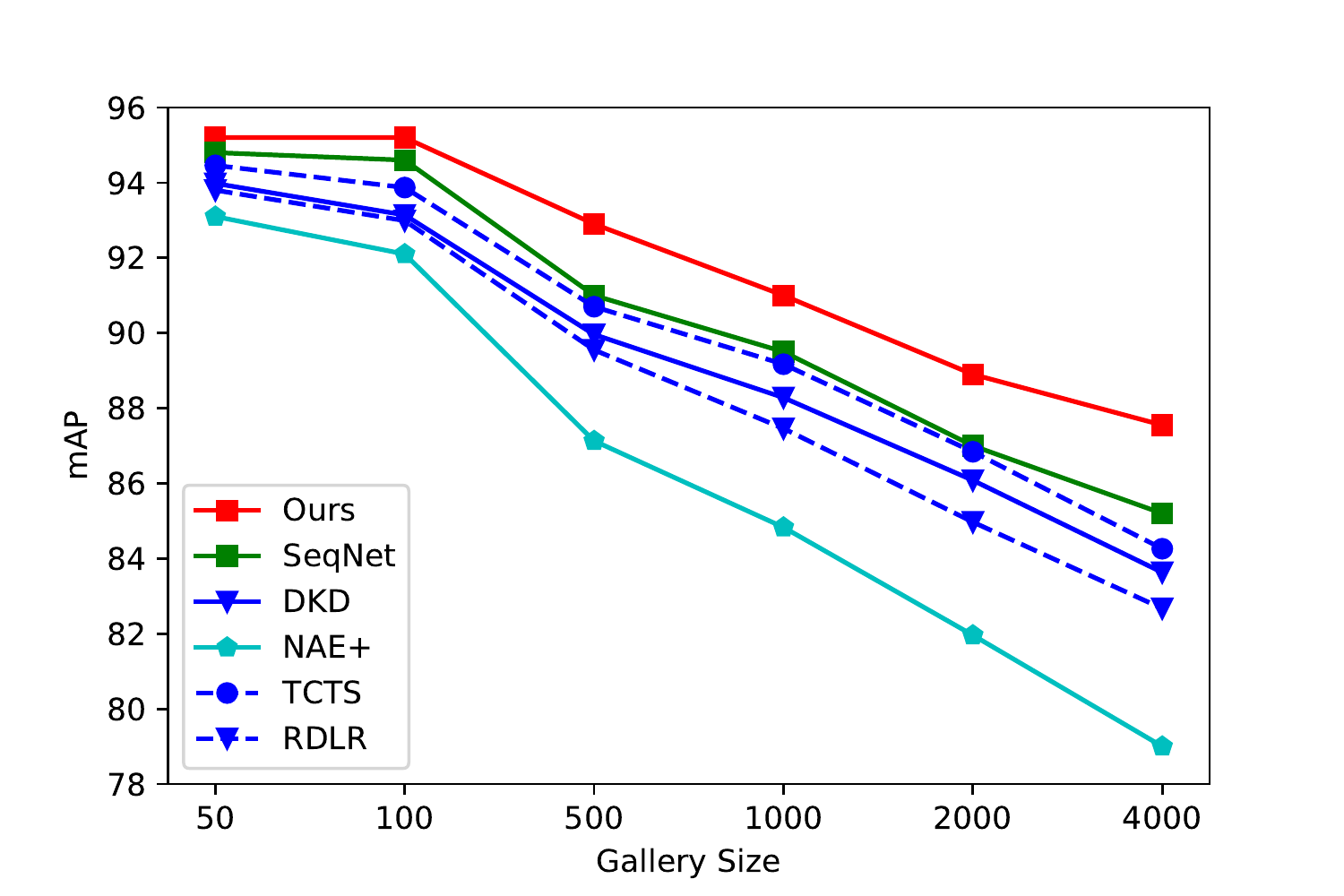}
  \end{center}
  \caption{State-of-the-art comparison of existing methods over CUHK-SYSU dataset with varying gallery sizes. Dotted lines represent two-stage methods whereas solid lines represent one-stage methods. Our PS-ARM shows consistent improvement compared to other methods as the size of gallery increases.  }
   \label{fig:gallery}
\end{wrapfigure}

To further analyse the benefits of our  ARM module, we introduced the proposed ARM module in to a Faster R-CNN based method  (NAE \cite{chen2020norm} method) after RoIAlign pooling.   We observed  that our ARM module can provide an absolute  gains of 1.9\%  and 1.8\% to the mAP  and top-1 accuracies  over NAE (see Table  \ref{tbl:quantitative}). Our PS-ARM outperforms all existing methods, and achieves  a mAP score of 95.2.  In terms of top-1 accuracy our method sets a  state-of-the-art accuracy of 96.1\%.

\begin{figure}[t!]
  \begin{center}
    \includegraphics[width=\textwidth]{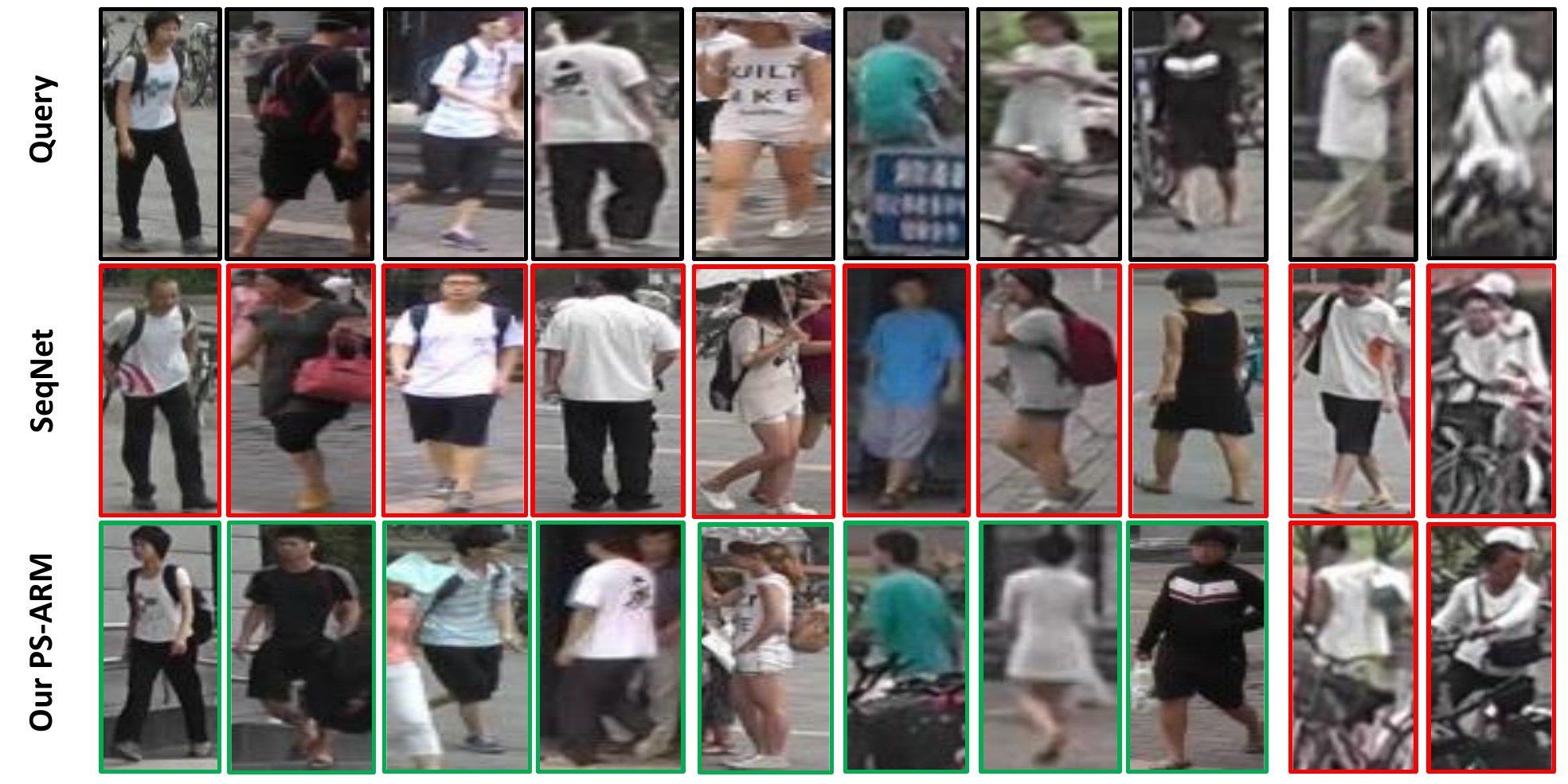} 
    \caption{ {Qualitative comparison between the top-1 results obtained from SeqNet (row 2)  and our PS-ARM (row3)  for the same query input (row 1 ).   Here, true and false matching results
are marked in green and red, respectively. SeqNet provides  inaccurate predictions due to the appearance deformations in these examples whereas our  PS-ARM provides accurate predictions by explicitly capturing  discriminative relation features within RoI.}}
\label{qualitative_sota_baseline_prw_dataset}
\end{center}
\vspace{-0.5cm}
\end{figure}

\begin{figure}[t!]
  \begin{center}
    \includegraphics[width=\textwidth]{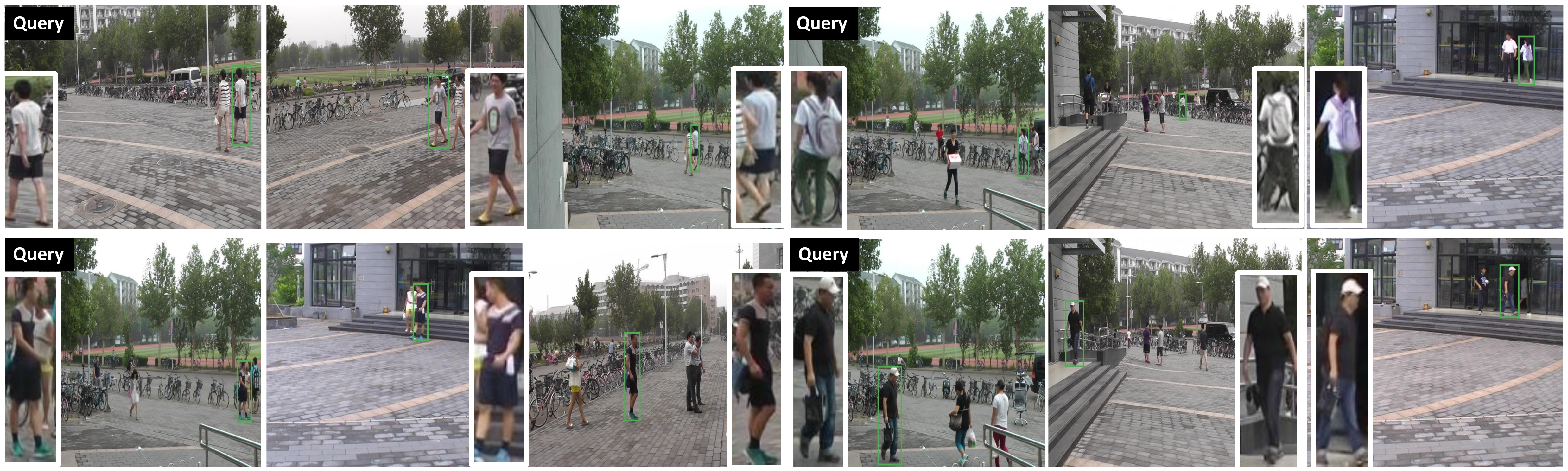}
    \caption{Qualitative results of our PS-ARM on challenging PRW dataset. The top-2 matching results for  each query image is shown. Our PS-ARM accurately detect and re-identify the query person in both  images.
    }\label{qualitative_prw_dataset}
\end{center}
\end{figure}

CUHK-SYSU dataset has different range of gallery sizes such as 50, 100, 500, 1000, 2000, and 4000.
To further analyze our proposed method, we performed an experiment  by varying the gallery size. Our mAP scores across different gallery size are compared with recent  one-stage and two-stage methods as shown in Figure~\ref{fig:gallery}. The results  shows that  our PS-ARM provides consistent performance gain over other approaches across all gallery sizes. 

\noindent\textbf{PRW Comparison:}
Table \ref{tbl:quantitative} shows the state-of-the-art comparison on PRW dataset. Among the existing two stage methods, MGN+OR \cite{yao2020joint} achieves the best  mAP  score 52.3, but with a very low top-1 accuracy. While comparing the top-1 accuracy, TCTS \cite{wang2020tcts} provides the best performance, but with a very low mAP score.  To summarize, the performance of most two-step methods \cite{yao2020joint,chen2018person,han2019re,girshick2015deformable,dong2020instance,lan2018person} are inferior either in mAP score or top-1 accuracy.

\begin{wrapfigure}{r}{0.5\textwidth}
  \begin{center}
    \includegraphics[width=0.48\textwidth]{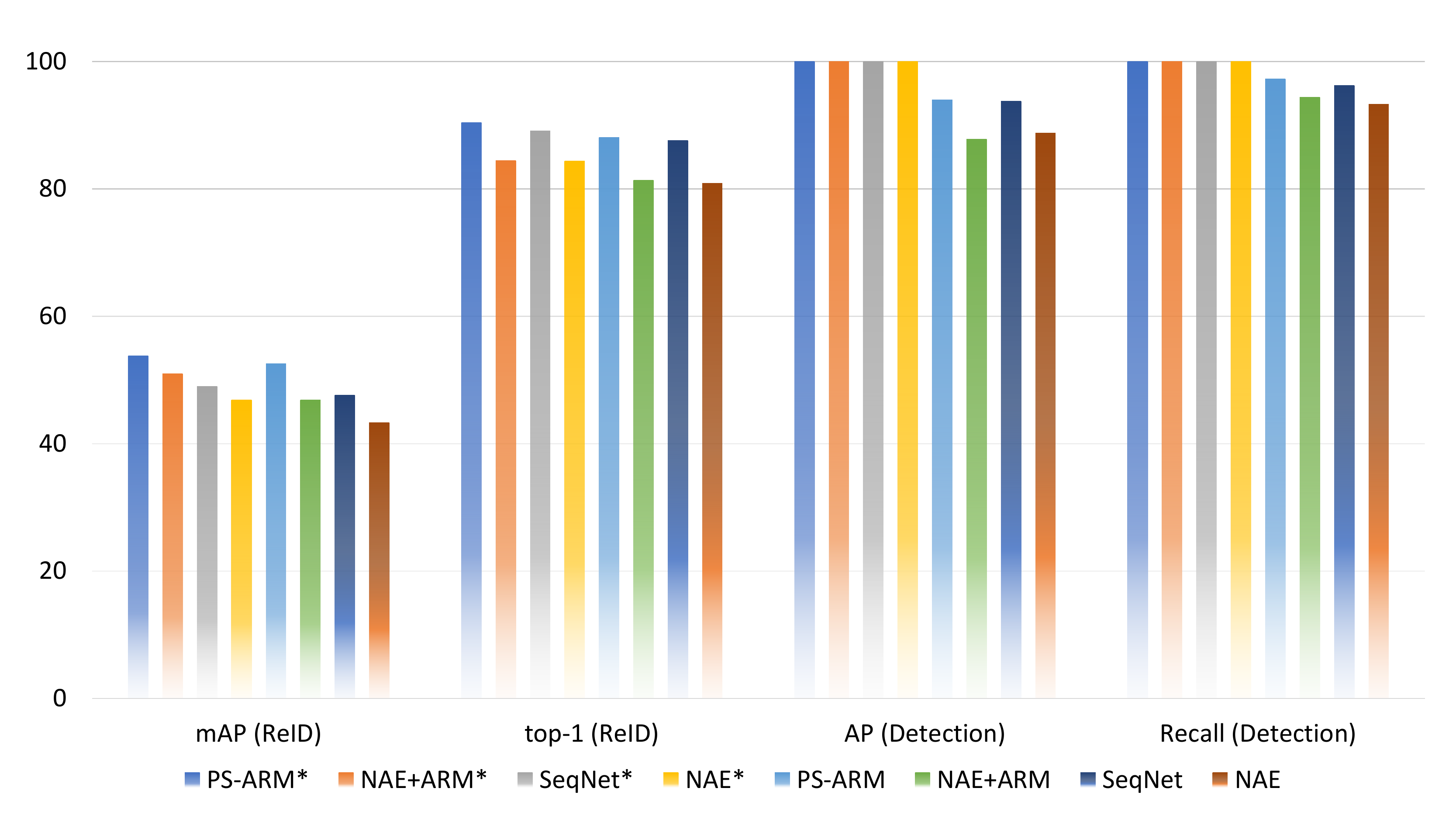}
  \end{center}
  \caption{{Person search and detection scores on PRW dataset with and without provided ground-truth detection boxes. The $*$ indicates the results using ground-truth boxes.}}
   \label{fig:detection_scores}
\end{wrapfigure}

Among  one-stage methods, NAE \cite{chen2020norm} and AlignPS \cite{yan2021anchor}, achieved mAP scores of  43.3\% and 45.9\%. These methods achieved  top-1 accuracies of  80.9\% and 81.9\%. Among the other one-step methods  SeqNet \cite{li2021sequential}, PBNet \cite{tian2020end}, DMRN \cite{han2021decoupled}, and DKD \cite{zhang2021diverse} also performed well and obtain more than 46\% mAP and have more than 86\% top-1 accuracy. 

To further analyze the effectiveness of our ARM module, we integrate our ARM module to NAE and achieved absolute mAP gain of  3.6\% mAP, leading to an mAP score of 46.9\%. We observe a similar performance gain  over top-1 accuracy,  resulting in  top-1 score of  81.4\%.  We  also introduced proposed ARM module in Han's \cite{han2021end} method. Compared to the existing methods, \cite{han2021end} utilize a different  approach, such as an RoI pooling of $24\times8$ size, instead of $14\times 14$.   To this end, we modified our PS-ARM to adapt the setting of \cite{han2021end}, resulting in an absolute gains of   2\% and 1.3\% improvement on PRW dataset and obtained 55.3\% mAP and 89.0\% top-1 scores, respectively.
Our PS-ARM  achieve state-of-the-art performance  compared  the existing one-step and two-step  methods. We achieve an  mAP score of 52.6\% and  top-1 score of 88.1\%.

 {Besides. similar to cascade  RCNN \cite{cai2018cascade}, we extend our person search network by introducing an other person re-id branch, called Cascaded PS-ARM. This newly introduced branch takes refined bounding boxes from the Box2 as an input to perform RoIAlign pooling. This strategy further refines the detection and re-identification, producing improved mAP 53.1\% and top-1 88.3 \% scores.}
 
\noindent \textbf{Qualitative comparison:}
Figure.~\ref{qualitative_sota_baseline_prw_dataset} shows qualitative  comparison between the  SeqNet \cite{li2021sequential} (row 2)  and our PS-ARM for the same query input (row 1). Here, true and false matching results are marked in green and red, respectively.  The figure shows top-1 results obtained from both methods. It can be observed that  SeqNet provides  inaccurate top-1 predictions due to the appearance deformations. Our  PS-ARM provides accurate  predictions on these challenging examples  by explicitly capturing  discriminative relation features within RoI. 
Figure.~\ref{qualitative_prw_dataset} shows the qualitative results from our PS-ARM. Here we show the top-2 matching results for each query image. It can be seen that our PS-ARM can accurately detect and re-identify the query person in both gallery images.

\begin{table}[t!]
\begin{center}
\caption{{Ablation study over the PRW dataset by incrementally adding our novel contributions to the baseline.  While introducing  a  MLP mixer to a baseline, both the detection and re-id performance increases over the baseline except top-1.   The spatially attended spatial mixing and channel-wise attended channel mixing within our relation mixer captures discriminative relation features within RoI while suppressing distracting background features, hence provides superior re-id performance.  Finally, our joint spatio-channel attention removes distracting backgrounds in a joint spatio-channel space, leading to improved detection and re-id performance. }}
\label{tbl:ablation_study}
\resizebox{\textwidth}{!}{
\begin{tabular}{|l|cc|cc|}
\hline
\multirow{2}{*}{Method} & \multicolumn{2}{c|}{ReID} & \multicolumn{2}{c|}{Detection} \\ \cline{2-5}
                        & mAP          & top-1          & Recall         &  AP      \\\cline{1-5} \hline \hline
Baseline & \multicolumn{1}{c|}{47.6}   & 87.6 & \multicolumn{1}{c|}{96.3}    & 93.1  \\ 
Baseline + MLP-Mixer  & \multicolumn{1}{c|}{49.1} & 86.8 & \multicolumn{1}{c|}{96.3} & 93.3   \\
Baseline + Transformer  & \multicolumn{1}{c|}{47.9}   & 85.8 & \multicolumn{1}{c|}{96.1} & 93.5  \\
Baseline + Spatio-channel Attention Layer  & \multicolumn{1}{c|}{48.1}   & 86.2 & \multicolumn{1}{c|}{95.2} & 93.0  \\
Baseline + Spatial Mixing + Channel-wise Attended Channel Mixing  & \multicolumn{1}{c|}{49.4} & 86.7 & \multicolumn{1}{c|}{96.2} & 93.2   \\
Baseline + Spatially Attended Spatial Mixing + Channel Mixing & \multicolumn{1}{c|}{49.5} & 86.9 & \multicolumn{1}{c|}{96.5} & 93.4   \\
Baseline + Relation Mixer  & \multicolumn{1}{c|}{51.8}   & 87.9 & \multicolumn{1}{c|}{96.6} & 93.8  \\
PS-ARM (Baseline + ARM)  & \multicolumn{1}{c|}{52.6} & 88.1 & \multicolumn{1}{c|}{97.1} & 93.9  \\
\hline
\end{tabular}}
\end{center}
\end{table}

\subsection{Ablation study}
Here, we perform the ablation study on the PRW dataset. Table.~\ref{tbl:ablation_study} shows the performance gain obtained by progressively integrating our novel contributions to the baseline.
 {
First we verify the effectiveness of the context aggregators including MLP-Mixer \cite{tolstikhin2021mlp} and Transformer \cite{Dosovitskiy_ViT_ICLR_2020} within the proposed framework.  The experiment shows that choice of MLP-mixer is better.
{Moreover, we apply joint spatio-channel attention on the RoI feature maps which results in improved performance compared to baseline. }
Further, we investigate the  introduction of spatially-attended spatial mixing and channel-wise attended channel mixing within our relation mixer which captures discriminative relation features within RoI while suppressing distracting background features.  This resulted in superior re-id performance. Introducing our relation mixer comprising of a spatially attended spatial mixing and channel-wise attended channel mixing  leads to an overall AP of 93.8 for detection and 51.8 mAP for re-id. } To further complement  the relation mixer that performs information mixing in the disjoint spatial and channel spaces, we introduce a joint spatio-channel attention.  Our joint spatio-channel attention removes distracting backgrounds in a joint spatio-channel space, leading to improved detection and re-id performance by achieving 94.1 and 52.6, respectively.

\subsubsection{Relation between Detection and ReID}
 {In Figure \ref{fig:detection_scores}, we validate the effectiveness of the proposed PS-ARM to deal with the contradictory detection and ReID objectives. We compared our PS-ARM with the SOTA SeqNet \cite{li2021sequential} and NAE \cite{chen2020norm}. We notice that PS-ARM$^*$ and NAE+ARM$^*$ outperforms their counter parts provided the ground-truth boxes.}

\section{Conclusions}
We propose a novel person search method named  PS-ARM, that strives to capture  global relation between different local regions within RoI of a person. { The focus of our design is introduction of a novel ARM module, which effectively capturing  the global  relation within an RoI and make robust against occlusion.  The relation mixer block introduces a  spatially attended spatial mixing, a channel-wise attended  channel mixing, and an input-output feature re-using for capturing  discriminative relation features within an RoI.  An additional  spatio-channel attention layer is introduced within the ARM module to further enrich the discriminability between the foreground/background  features  in a joint spatio-channel space}. Our ARM  module  is  generic  and  it can be easily integrated to  any  Faster  R-CNN based person search methods.  Comprehensive experiments are performed on two benchmark datasets.  We achieve state-of-the-art performance on both datasets,  demonstrating the merits  of our novel contributions.


%
%
%

\bibliographystyle{splncs04}
\bibliography{mybibliography}
\end{document}